\documentclass{article}
\usepackage[T1]{fontenc}
\usepackage[utf8]{inputenc}
\usepackage{caption}
\usepackage{amsmath}
\usepackage{amssymb}
\usepackage{mathrsfs}
\usepackage{graphicx}
\usepackage{ntheorem}
\usepackage{algorithm}
\usepackage[noend]{algpseudocode}
\usepackage[sfdefault=cmbr,OMLmathsans]{isomath}  % notation acc. to ISO 31 / 80000
\usepackage{upgreek}  % upright greek letters
\usepackage{tensor}  % for tensor index notation
\usepackage{mathtools}
\usepackage{caption}
\usepackage{subcaption}
\usepackage[table,x11names]{xcolor}

\newcommand{\ignore}[1]{}

\newcommand{\vek}[1]{\mathchoice{\displaystyle\boldsymbol{#1}}
{\textstyle\boldsymbol{#1}}{\scriptstyle\boldsymbol{#1}}
{\scriptscriptstyle\boldsymbol{#1}}}

\def\realnr{\mathbb{R}}

\providecommand{\keywords}[1]
{
  \small	
  \textbf{\textit{Keywords---}} #1
}

\usepackage[style=alphabetic]{biblatex} 
\begin{document}
%\title{Finding  laws valid inside a  Data Set and classifying it  by those laws  using Neural Networks}
\title{Finding hidden-feature depending laws inside a data set and classifying it using Neural Network.}
\author{Thilo Moshagen         \and
     Nihal Acharya Adde \and
     Ajay Navilarekal Rajgopal}
\maketitle
\begin{abstract} 
The logcosh loss function for neural networks has been developed to combine the advantage of the absolute error loss function of not overweighting outliers with the advantage of the mean square error of continuous derivative near the mean, which makes the last phase of learning easier. It is clear, and one experiences it soon, that in the case of clustered data, an artificial neural network with logcosh loss learns the bigger cluster rather than the mean of the two. Even more so, the ANN, when used for regression of a set-valued function, will learn a value close to one of the choices, in other words, one branch of the set-valued function, while a mean-square-error NN will learn the value in between.  This work suggests a method that uses artificial neural networks with logcosh loss to find the branches of set-valued mappings in parameter-outcome sample sets and classifies the samples according to those branches.
\end{abstract}


\section{Introduction}
Given a set of data tuple, \emph{Clustering} algorithms \parencite{clustering} decide which elements of the set belong together, i.e. form a subset in the sense that they have closer mutual distance among each other. Further, there are a lot of well established and also new methods to deal with the question of whether two or more sets of samples belong to the same population or not. Mainly, this considers the field of statistical hypothesis testing \parencite{1999kats.book.....S} which is a testable hypothesis based on observed data modelled as the realised values taken by a collection of random variables. Also, in a different setting, a data model can be defined as a set of mathematical laws that might be valid inside a data set and describes how the data elements relate to one another. Given that there exist some measurements and parameters that caused these measurements, the model selection tells which model is most likely valid for the observed measurements to happen. Variants of ANOVA (Analysis of variance) combine the two and is used to analyse the differences among group means in a sample \parencite{doi:https://doi.org/10.1002/9781118445112.stat06938}.
\newline
\par
We consider a method that answers the question whether a set of vector-valued samples, where some components can be seen as cause and at least one as an effect,  obeys some possibly unknown rule, or if it rather splits into groups that fulfil different rules. In other words, assuming that any input data point consists of components $\vek x$ that are presumably independent and component(s) $\vek y$ that depend on them by some generally unknown rules, the suggested methods finds the rules in the shape of an artificial neural networks' weights \emph{and} clusters the data into groups obeying each of the found rules.
\newline
\par
The three key features of the suggested method are, first, the use of neural networks to extract one of the unknown rules that are valid in parts of the data. This extraction is done by supervised learning, which is a regression in mathematical terms. The second key feature is that supervised learning is done with a loss function that is approximately linear in the distance to zero and thus puts less weight on far-off data than the square error loss function. For example, the $L_1$-norm fulfils this. But here, the logcosh loss function was used as it facilitates learning, while still having the desired property. When used for regression, such a loss function leads to learning a function that approximates well the strongest cluster of output data, while hardly taking into account clusters with fewer members. Data lying away from the found regression graph thus is probably obeying another law; Distance to the regression function found by the artificial neural network is then used as a classification criterion. This is the third key feature. The data that is approximated well by the found regression function is considered to be governed by that function. With the badly approximated data, a new network is taught, and all points where its forecast matches are considered to be governed by that second regression. This procedure is continued until no relevant data remains unclassified.   This is how in brief the suggested method works.

\section{Problem Setting} \label{section:ProblemSetting}
\subsection{Mathematical Description}
%\label{section:1}
Let 
\begin{equation}
\left\{ (\vek x, \vek y) \in\realnr^d \times \realnr^{d} \right\}
\end{equation}
be a set of data points, where it can be assumed that $\vek y$  depends on $\vek x$. % given by measurement
To simplify the setting and also due to the fact that artificial neural networks do not encourage vector valued output, 
we restrict ourselves to 
\begin{equation}
%X = 
\left\{ (\vek x,  y) \in  \Omega \subset  \realnr^d \times \realnr \right\}
\end{equation}
%provided that close-by $\vek x$ frequently have distant $\vek y$. 
%Let further   
%\begin{equation}
%X\subset \Omega \subset \realnr^d \times \realnr 
%\end{equation}
%be a subset that contains the data points
where $\Omega$ is the domain in which observations are defined.
%where the function 
%\begin{align}
%\label{1}
%\Phi_i: X & \longrightarrow  \realnr \\ \label{2}
%\vek x &\mapsto  \Phi_i(\vek x) =  Y(\vek x)
%\end{align}
%is defined and $\Phi$ is the function that maps $X$ and $Y$. 
%The function $\Phi$ is such that one parameter $\vek x$ might have multiple outputs $\vek y$.
\newline
\par

%Mine, new: 
The presence of clusters $\left\{(\vek x, y)\right\}_{i, i=1,...,M}$, where inside each 
cluster, %$\left\{(\vek x, y)\right\}_{i, i=1,...,M}$  
the $(\vek x, y)$-tuples obey a different law is now mathematically described as follows:
\\
\par
Each clusters' independent variable points are subsumed in the set $\Hat X_i\subset X $, $X$ being all $\vek x $ in  $ \Omega  $,  where the law valid inside it is (first defined on the samples only, with the hat denoting this): 
% Now we consider  

\begin{align}
%X_i\subset& X, \quad i = 1,...,N<\infty\\
\Hat \phi_i: \Hat  X_i & \longrightarrow  \realnr \\
\vek x &\mapsto \Hat  \phi_i(\vek x) =  y(\vek x)
%\vek \phi:\realnr^d &\longrightarrow & \realnr \\
\end{align}
where $\Hat \Phi_i$ is a single-valued function, mapping each point in $\hat X_i$ to a unique value in the range. The existence of multiple $\Hat \Phi_i$ is due to hidden features, for which nearby $\vek x$ can have very distant $y$.
Each $\Hat \Phi_i$ induces a continuous function $\Phi_i$ in some super-set  $X_i$ of $\Hat X_i$ by regression, the continuous regression counterpart of the measurements. Those then have reasonably bounded derivatives - which a mapping $\Phi$ that maps all $x$ would not have.
There may exist a certain subset of $X$ which gives the same output %$\{Y\}$ 
 for all  $\Phi_i$, while in the $X_i$ the $\Phi_i$  give different values.
% Now, the data is such that, one parameter $\vek x$ may have more than one image under $\Hat{\phi}$ -- 
% some $\vek x$ may have distant $\vek y$ in spite of being very close to each other. If looking for continuous functions $\Phi$, this means that one looks for a possibly  \emph{set-valued} function: 
%The nomenclatura is chosen to express this: 
%\begin{align}
%\Hat X_i\subset& X, \quad i = 1,...,M<\infty\\
%%\Omega_i\subset& \Omega, \quad i = 1,...,N<\infty\\
%\Hat{\phi}_i:%\Omega_i 
%\Hat X_i & \longrightarrow  \realnr \\
%\vek x &\mapsto \Hat{\phi}_1(\vek x) = y_1(\vek x)\\
%\vek x &\mapsto \Hat{\phi}_2(\vek x) = y_2(\vek x)
%%\vek \phi:\realnr^d &\longrightarrow & \realnr \\
%\end{align}
%where each component of $\Hat{\phi}$  represents one of the possible outcomes which are not distinguishable by $\vek x$ a priori, for  which, a rule is valid depending on some hidden property: 
% It is known in the beginning which data belongs to which situation, or the different rules for the different outcomes of $\hat{\phi}(\vek x)$. We assume that the entries of $\hat{\phi}$ exist and possess some smoothness. They 
%
Thus, $\Phi_i$ may be seen as defined only on  $X_i$, or alternatively on $X$, in which case the $\Phi_i$ coincide in parts of $\Omega$. 
This  can be  seen as a multi-valued or set-valued function %gives a multi-valued output. 
\begin{align}
\Phi (\vek x) = \left\{ \begin{array}{cc} 
                \Phi_1 \\
                \vdots\\
                \Phi_M
                %\Hat{\phi}\\
                \end{array} \right\}.
\end{align}The set-valuedness in this nomenclature is expressed by this vector-valuedness.  It captures the property that the data input-output pairs indeed belong to different situations or populations. 
%\par
The task to solve in this nomenclature is: Given the set $X$, find the rules $ \Phi_i$ and the subsets $X_i$ where they are valid.
\newline
\subsection{Outline of Strategy}
One seeks to learn each $X_i$'s rule $\Phi_i$ by regression, which for general $\Phi_i$ is done best 
%The discussed problem setting %creates
%is a  data set
%which %can be otherwise described as clustered
%is clustered by the multi-valuedness ... As for some $\vek x$ there are 2 or more possible $y$, the 
 by artificial neural network,
%trained with this as an input considers it as clustered data. %The logcosh loss function can be used to train such a setting to classify the bigger cluster. 
using the  logcosh loss function: It weights the outliers less, similar to the MAE loss while it exhibits good performance during gradient descent as MSE. The network trained with logcosh loss % tries to
will thus learn the biggest cluster $\Phi_1$ efficiently because it weights smaller clusters away from the biggest one only linearly with distance, unlike the squared error losses, % without being affected by the smaller clusters
and thus classifies the data as belonging to the biggest cluster or not. 
In our research, we %??give different weights to the clusters and 
 train the network with logcosh loss function in an aim to classify the clustered data.  This approach is demonstrated using a simple 1-dimensional and 2-dimensional problem.

 \section{ Artificial Neural Network Regression Quality  as a Classification Criterion}
Supervised learning of an Artificial Neural Network \parencite{Goodfellow-et-al-2016} has the task of learning a function that maps an input to an output based on example input-output pairs. It is where the set of input variables $\vek x$ and the output variables $\vek y$ are available and one has to use an algorithm to learn the mapping function from the input to the output. The goal is to approximate the mapping function so well that the new unseen input data $\vek x$ can be used to predict the output variables $\vek y$ for that data. An ANN is based on a collection of connected nodes called neurons which loosely represents the neurons in a biological brain. Each connection transmits signals from one neuron to the other. The signal at a connection is a real number, and the output of each neuron is computed by some non-linear function of the sum of its inputs. These connections are called edges. Neurons and edges typically have a weight that adjusts as learning proceeds. Through backpropagation, the network tries to find optimal weights and biases to represent the model. In other words, the artificial neural network can be represented as an optimization problem which ultimately is equivalent to minimising the loss function of the data. Therefore the choice of the loss function becomes vital for modelling an efficient network. Our task is to find an appropriate model that fits the regression model by one of the rules and classifies the clustered data by it. Therefore, in the following section,  we discuss the different available loss functions and choose an appropriate loss function in an aim to classify the clustered data.

\subsection{Loss Functions Properties}
One key feature of the suggested method is the choice of the loss function. We will point out in the following that for a regression problem, the minimizer of loss functions that rise linearly with the distance lies inside a cluster, while for the quadratic loss functions, it lies between clusters. The choice of loss function depends on a number of factors including the presence of outliers, choice of the machine learning algorithm, time efficiency of gradient descent, ease of finding the derivatives and confidence of predictions. \parencite{loss} investigated some representative loss functions and analysed the latent properties of them. The main goal of the investigation was to ﬁnd the reason why bilateral loss functions are more suitable for regression task, while unilateral loss functions are more suitable for classiﬁcation task. This section covers in detail the different loss functions which can be used for our regression problem as discussed by \parencite{lossfuncations}.

\subsubsection{Mean Square Error (MSE) or L2 loss}
This function originates from the theory of regression, least-squares method. Mean Square Error (MSE) is the most commonly used regression loss function. MSE is the sum of squared distances between our target variable $y$ and predicted values $y_{p}$.
\begin{equation}
\mathbf{M.S.E.} = \frac{\sum_{i=1}^{n} (y^{i}-y_{p}^{i})^{2}}{n}
\end{equation}
It is well known that, here, few distant points outweigh the closer points. % Was ich nicht sag, brauch ich nicht beweisen... MSE is sensitive towards outliers and given several samples with close input feature values, the optimal prediction will be the mean of their target values.  
The MSE loss establishes   %is great for ensuring 
that our trained model takes outliers seriously as the contribution to loss by an outlier in input is magnified by squaring and so learning results are biased in favor of the outliers. This can be an advantage -  predictions in zones with outliers do not produce  huge errors to the outliers since the MSE took them into account. MSE is thus good to use if the target data conditioned on the input is normally distributed around a mean value and in the absence of outliers. %, and when it is important to penalize outliers extra much. 
It has a continuous derivative and therefore the minimisation with gradient methods works well.
%The main disadvantage of MSE is when the model makes a single very bad prediction which would magnify the error due to squaring. 
The described property is a disadvantage for our setting, as one cluster consist of outliers seen from the other clusters' perspective, thus the MSE minimiser would be right in between clusters.
\\
\par
\begin{figure}[ht]
    \centering
    \includegraphics[width=8cm]{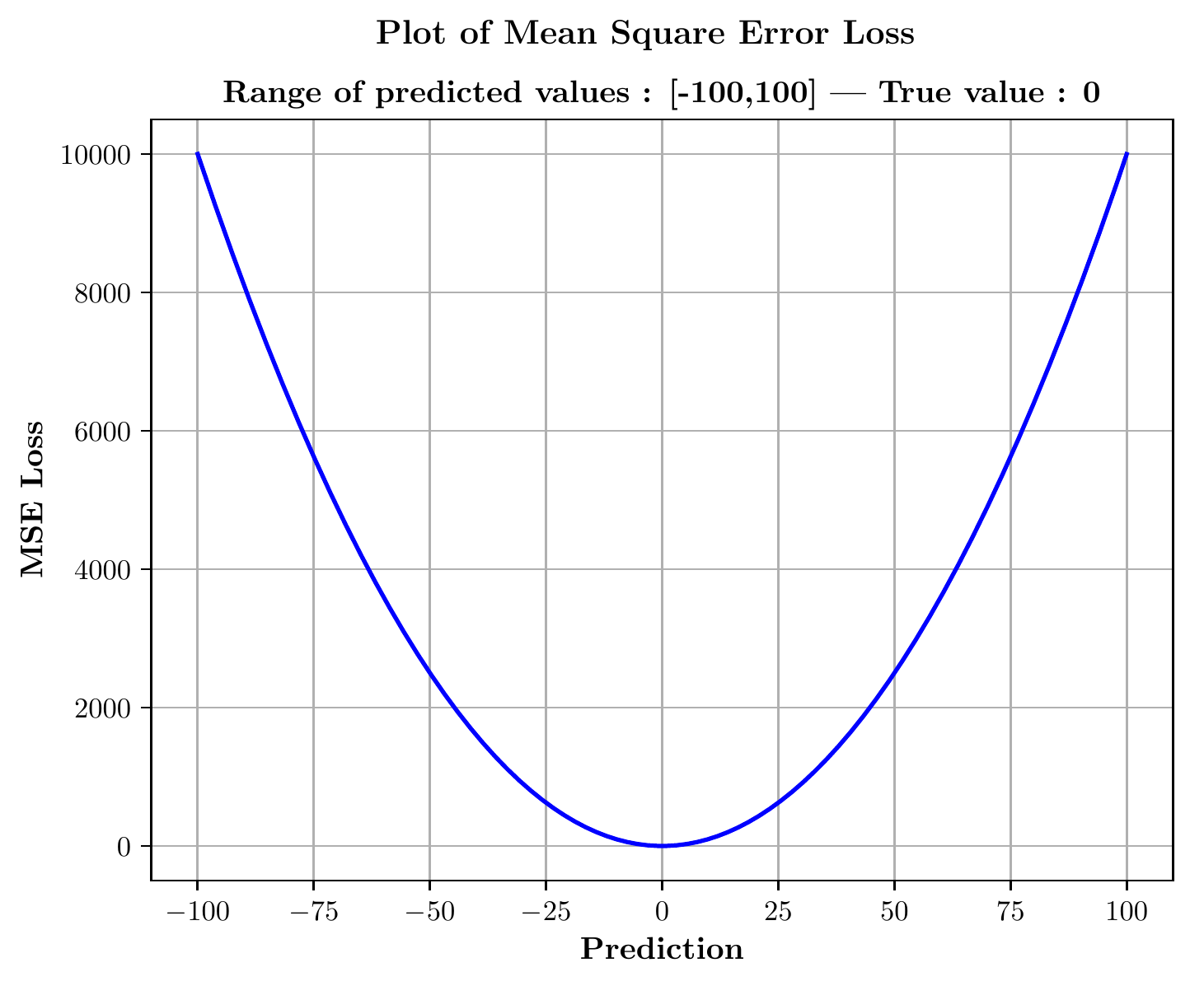}
    \caption{Plot of Mean Square Error (MSE) Loss}
    \label{fig:MSE}
\end{figure}

Figure \ref{fig:MSE} shows the plots of mean square error loss vs. predictions, where the target value is 0, and the predicted values range
between -100 to 100. The loss (Y-axis) reaches its minimum value at the prediction (X-axis) = 0. The range of the loss is 0 to $\infty$.

\subsubsection{Mean Absolute Error (MAE) or L1 loss}
Mean Absolute Error (MAE) is just the mean of absolute errors between the actual value $y$ and the value predicted $y_{p}$.  So it measures the average magnitude of errors in a set of predictions, without considering their directions.
\begin{equation}
\mathbf{M.A.E.} = \frac{\sum_{i=1}^{n}|(y^{i}-y_{p}^{i})|}{n}
\end{equation}

\begin{figure}[ht!]
    \centering
    \includegraphics[width=8cm]{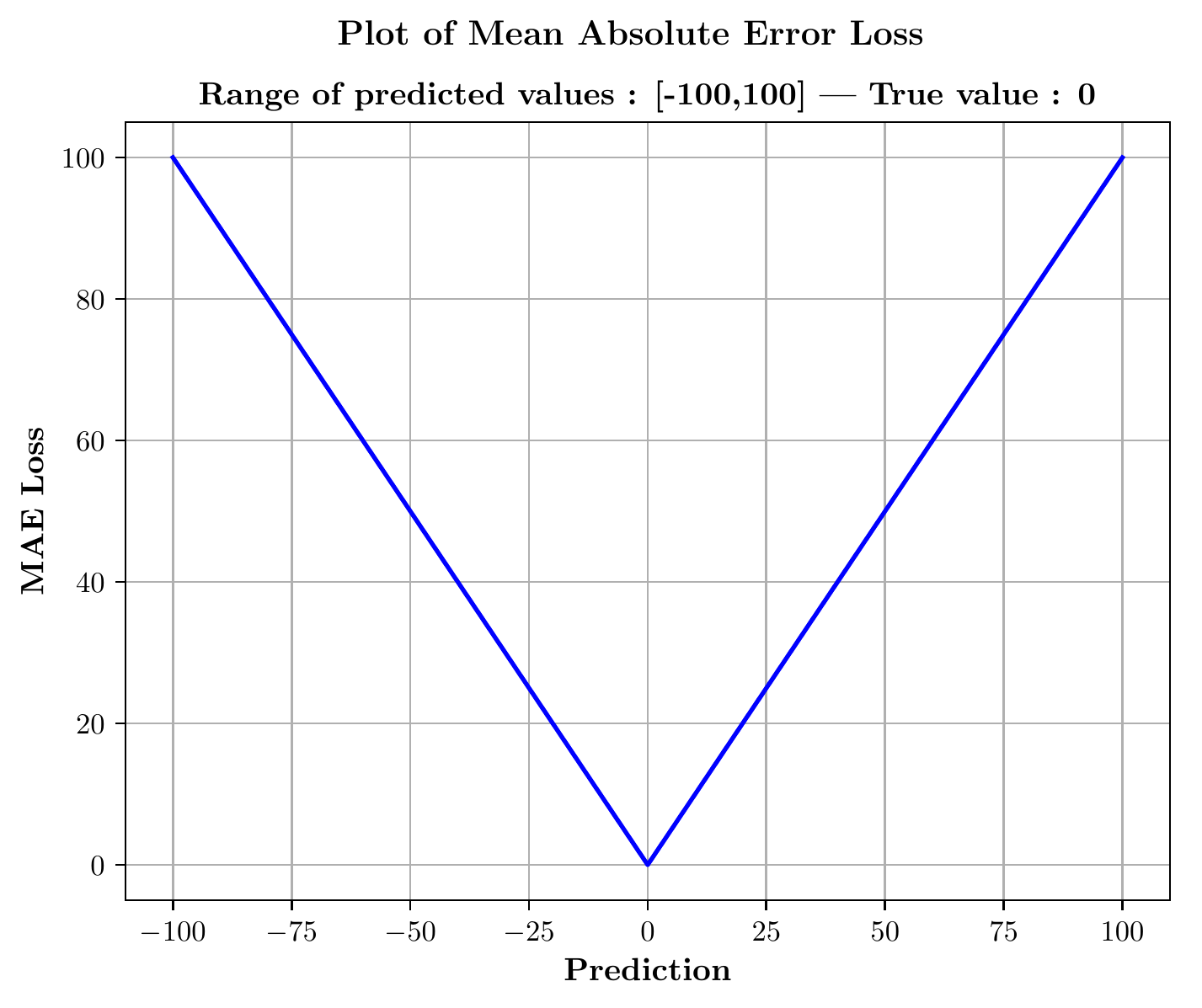}
    \caption{Plot of Mean Absolute Error (MAE) Loss}
    \label{fig:MAE}
\end{figure}

As one can see, for this loss function, both the big and small distances contribute the same. The advantage of MAE covers the disadvantage of MSE. As we consider the absolute value, the errors will be weighted on the same linear scale. Therefore, unlike the previous case, MAE doesn't put too much weight on the outliers. % and the loss function provides a generic and even measure of how well our model is performing. 
However, it does not have a continuous derivative and thus frequently oscillates around a minimum during gradient descent. The MSE does a better job there as it has a continuous derivative and provides a stable solution. % when occasional outliers don't exist.
Figure \ref{fig:MAE} shows the plot of mean absolute error loss with respect to the prediction  while the target value is 0, similar to the previous case. 
%Das ist so nicht zielführend: Now, since our data has some outliers, we would not want our predictions to be biased towards these outliers (by using $\mathbf{M.S.E.}$), nor do we want to ignore the outliers (by using $\mathbf{M.A.E.}$). Hence, we need to use some other loss function for our problem.\\

\subsubsection{Huber loss}
\begin{figure}[ht!]
    \centering
    \includegraphics[width=8cm]{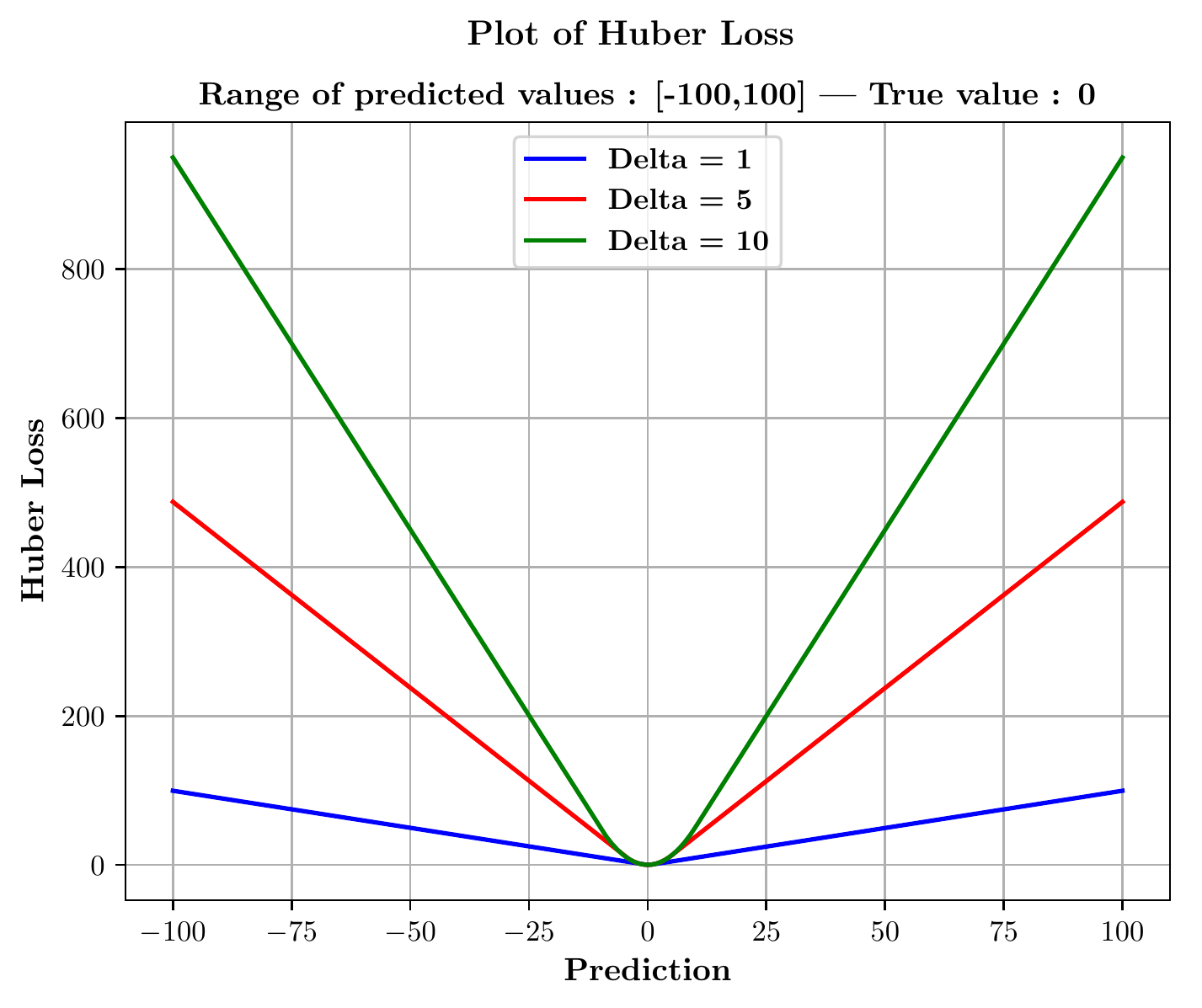}
    \caption{Plot of Huber Loss}
    \label{fig:Huber}
\end{figure}

Huber loss is just the absolute error but transforms to squared error for small values of error. It is an attempt to overcome MAE's disadvantage of nonsmooth derivative. Huber loss is less sensitive to outliers in data than the squared error loss. It is also differentiable at 0. It is basically absolute error, which becomes quadratic when the error is small. How small that error has to be to make it quadratic depends on a hyperparameter $\delta$, which can be tuned. Huber loss approaches MSE when $\delta \rightarrow 0 $ and MAE when $\delta\rightarrow \infty $ (large numbers). It is defined as 
\begin{equation}
L_{\delta}(y,y_{p}) =
\left\{
	\begin{array}{ll}
		\frac{1}{2}(y-y_{p})^{2}  & \mbox{if } |y-y_{p}| \leq \delta\\
		\delta|y-y_{p}|-\frac{1}{2}\delta^{2} & \mbox{otherwise } 	
	\end{array}
\right\}
\end{equation}
The choice of $\delta$ becomes increasingly important depending on what one considers as an outlier. Residuals larger than delta are minimized with L1 while residuals smaller than delta are minimized with L2. Hubber loss combines the advantages of both the loss functions. It can be really helpful in some cases, as it curves around the minima which decreases the gradient. However, the problem with Huber loss is that we might need to train hyperparameter delta which is an iterative process. Figure \ref{fig:Huber} shows the plot of Huber loss vs. predictions for different values of delta $\delta$.

\subsubsection{Log-Cosh loss}
\label{section:logcosh}
%Log-cosh is another loss function used in regression tasks which is smoother than L2. 
Log-cosh is the logarithm of the hyperbolic cosine of the prediction error. Given the actual value $y$ and the predicted value $y_{p}$, the log-cosh is defined as 
\begin{equation}
L(y,y_{p}) = \sum_{i=1}^{n}|\operatorname{log}(\cosh((y^{i}-y_{p}^{i})))|
\end{equation}

\begin{figure}[ht]
    \centering
    \includegraphics[width=8cm]{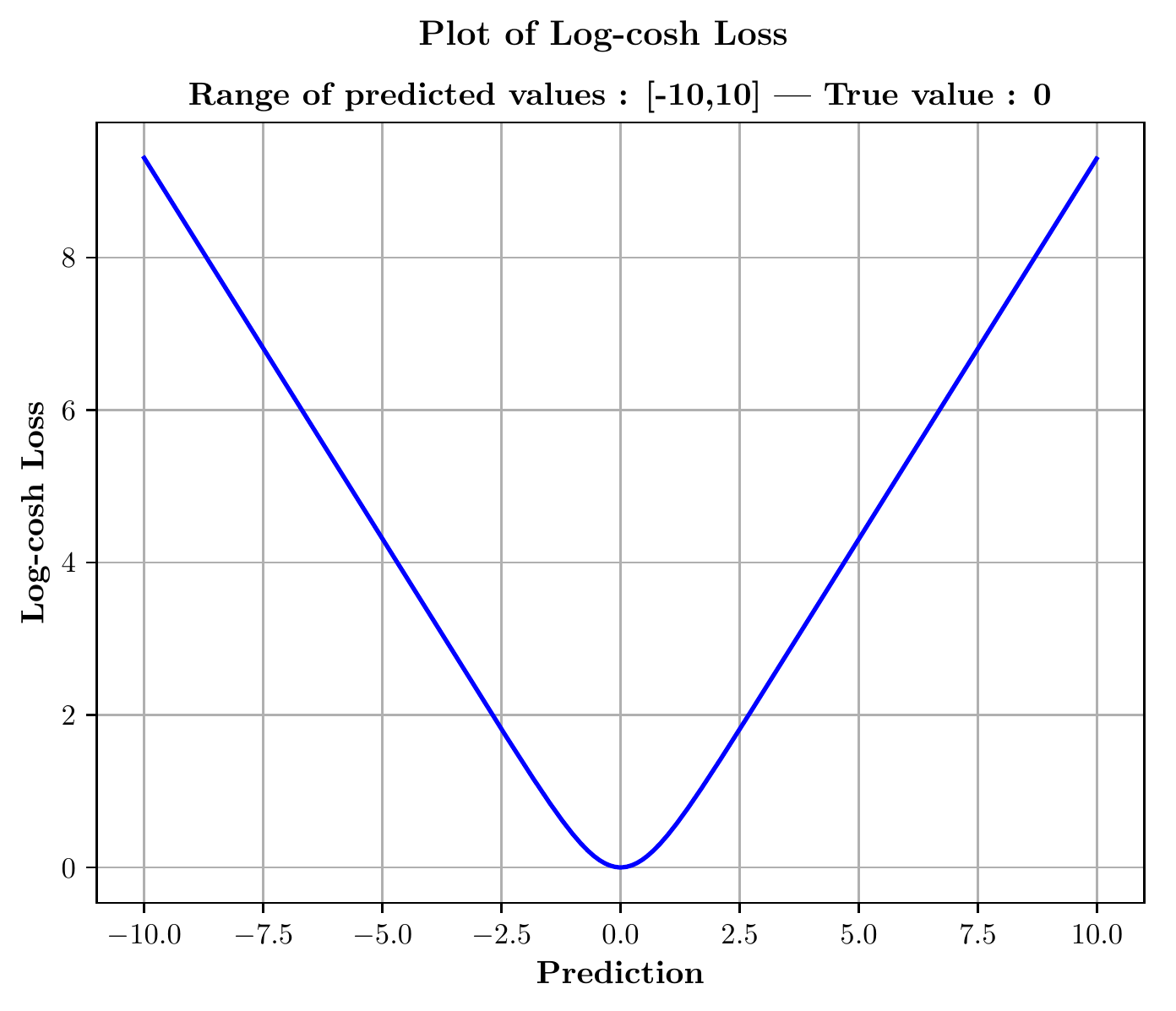}
    \caption{Plot of Log-cosh Loss}
    \label{fig:logcosh}
\end{figure}

$\operatorname{log}(\cosh(x))$ is approximately equal to $\frac{x^2}{2}$ for small values of x and to $|x|-\operatorname{log}(2)$ for larger values.
It is twice differentiable everywhere unlike Huber loss.  Therefore, the log-cosh loss function is similar to mean absolute  error with respect to its moderate weighting of outliers, while it behaves stable during gradient descent search. % not be largely affected by occasional wrong predictions. 
Figure \ref{fig:logcosh} shows the plots of logcosh loss vs predictions, where the target value is 0, and the predicted values range between -10 to 10. 
\newline
\par
%The logcosh loss function for neural networks has been developed to combine the advantage of the absolute error loss function of not overweighting outliers with the advantage of the mean square error of continuous derivative near the mean, which makes the last phase of learning easier.
Therefor, in our research log-cosh loss function was used and indeed showed  good results in classifying the data based on the hidden features.\\
\par
\begin{figure}[ht]
    \centering
    \includegraphics[width=8cm]{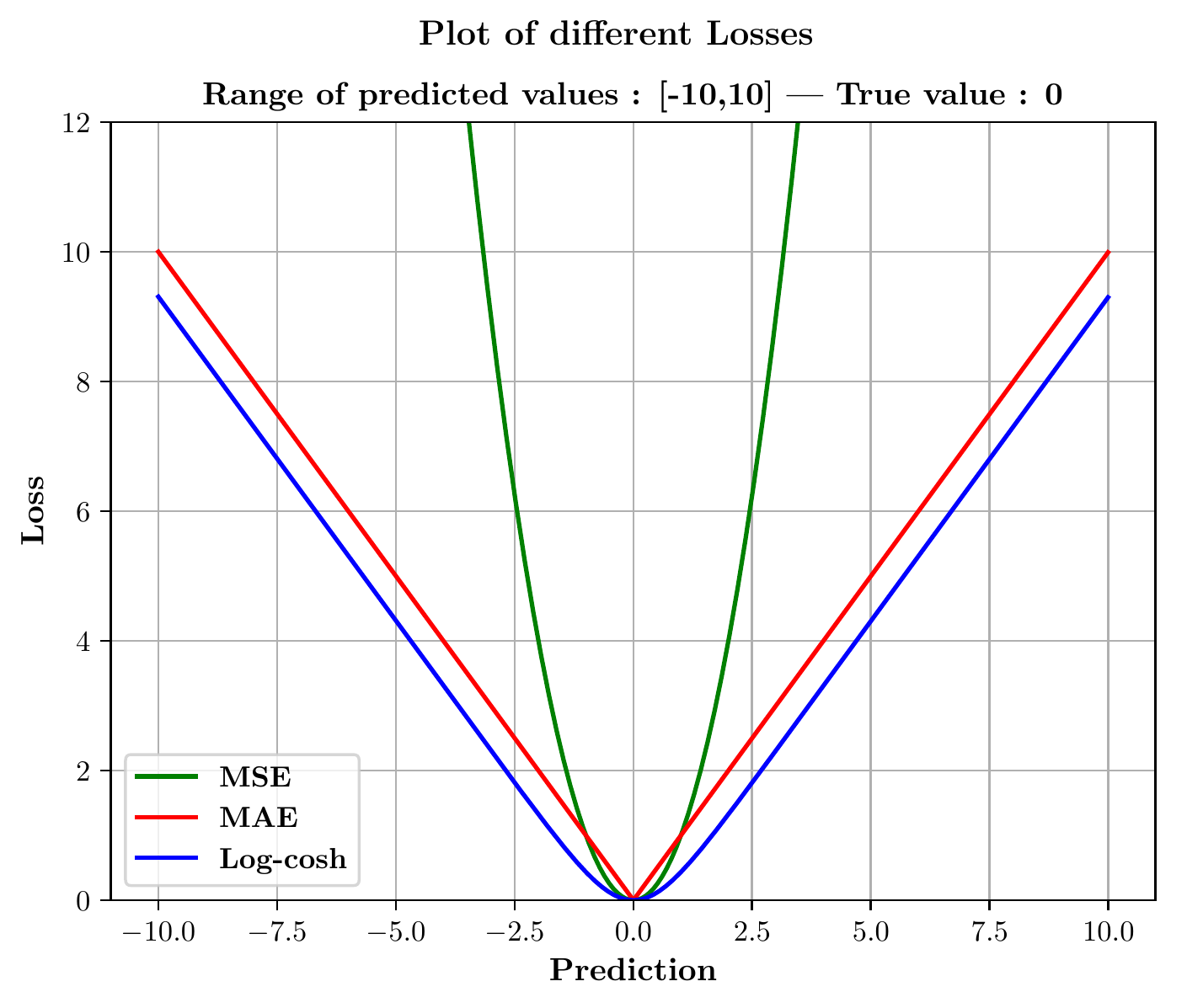}
    \caption{Plot of different Losses : MSE, MAE and Log-cosh}
    \label{fig:compareloss}
\end{figure}

\iffalse
ich denke, die Tabelle gibt nichts an, was man nicht in fig 5 sieht.
Table \ref{table:compare} gives example values of different loss functions for comparison: Mean absolute error, Root mean square error and Logcosh when occasional outliers are present. Here, RMSE (square root of MSE) is considered to consider it on the same scale as MAE. 
\newline
\begin{table}[ht]
\centering
\begin{tabular}{|c|c|c|c|}
\hline
\rowcolor{lightgray} \textbf{Error} & \textbf{|Error|}      & \textbf{(Error)\textasciicircum{}2} & \textbf{log(cosh(Error))} \\ \hline
0              & 0                     & 0                                   & 0                         \\ \hline
1              & 1                     & 1                                   & 0.434                     \\ \hline
2              & 2                     & 4                                   & 1.325                     \\ \hline
-2             & 2                     & 4                                   & 1.325                     \\ \hline
1.5            & 1.5                   & 2.25                                & 0.855                     \\ \hline
\rowcolor{yellow} 15             & 15                    & 225                                 & 14.307                    \\ \hline
               & \textbf{MAE :  3.583} & \textbf{RMSE: 6.275}                & \textbf{Log-cosh: 3.041}  \\ \hline
\end{tabular}
\caption{Comparison of MAE, RMSE and Log-cosh loss functions when outliers are present. }
\label{table:compare}
\end{table}
\fi
Figure \ref{fig:compareloss} compares the 3 different losses functions. 
When clustered data is present, an artificial neural network with logcosh loss function learns the bigger cluster rather than the mean of the two and hence can be used to classify the clustered data. In the case of MSE, due to the squaring of the error function, few faraway points are weighted more than the nearby points. When learning clustered data, the network with MSE loss function gets affected by these outlying clusters and tries to find the minima between them and thereby fails to learn the bigger cluster.  For linearly growing loss functions like logcosh and MAE, just the sum of distances counts and few far-away points do not count more than several nearby points and therefore, a regression value near or through the heavier cluster is learnt.  Though the MAE loss function has this property of  the bigger cluster, it is non-smooth and has a non-continuous derivative resulting in oscillating behaviour.  %As MAE is unstable during the last part of minimization, it oscillates between the clusters. 
As mentioned above, since the logcosh loss function is a combination of MAE for larger values and MSE for the smaller values, it successfully learns the bigger cluster and gives a stable solution. These features of logcosh loss function are exploited in our research.

\subsection{One-Dimensional Test Case}
\subsubsection{Test Problem}
We now consider a simple 1D example based on the concept discussed in the section~\ref{section:ProblemSetting}. Two simple single-valued polynomial functions were selected and combined in different fractions to achieve a multi-valued data set. This section discusses the problem setting of the 1-dimensional case and thereafter the network behaviour based on the chosen data set.
\\
\par
To create a multi valued Data set, 2 simple functions were selected as below.
\begin{equation}
\Phi_{1}(\vek x) = {((x-4)(x+4))}^2,\hspace{5mm} x \in [-6,6]
\end{equation}
\begin{align}
\Phi_{2}(\vek x) = \left\{ \begin{array}{cc} 
                {((x-4)(x+4))}^2 & \hspace{5mm} x \in [-6,-4) \\
                0 & \hspace{5mm} x \in [-4,4] \\
                {((x-4)(x+4))}^2 & \hspace{5mm} x \in (4,6] \\
                \end{array} \right\}.
\end{align}

where $\Phi_1$ and  $\Phi_2$ are two single-valued functions which are defined within the interval $[-6,6]$.

\subsubsection{Training Strategy}
The data set was split such that $80\%$ of the data were used for training and the rest $20\%$ were used as test data. Initially, both the functions were trained individually with a basic regression neural network and then tested on the test data to validate the network.
\newline
\begin{figure}[ht]
     \centering
     \begin{subfigure}[b]{.45\textwidth}
         \centering
         \includegraphics[width=\linewidth]{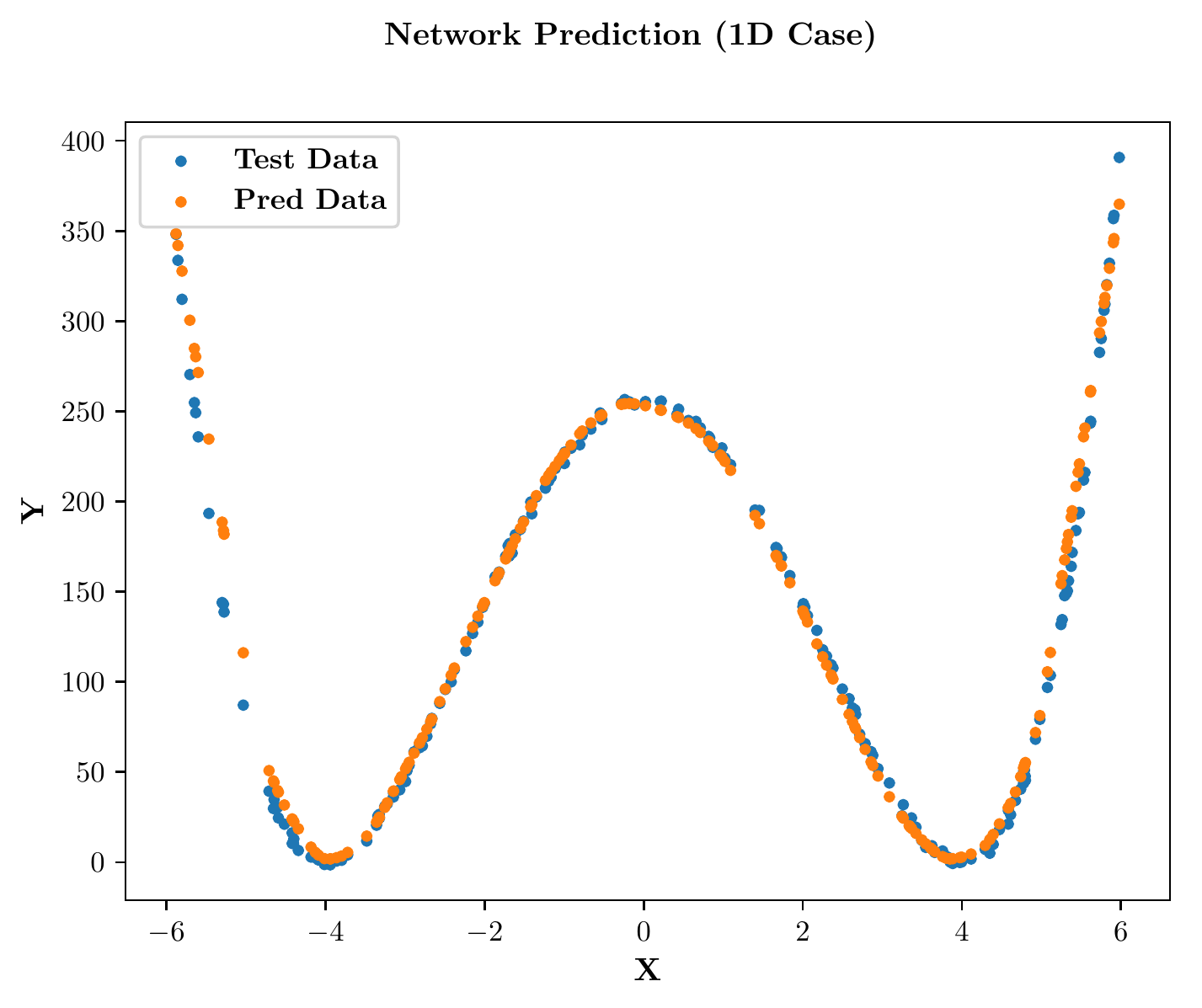}
         \caption{$60\%$ or more of function $\phi_1$}
     \end{subfigure}
     \begin{subfigure}[b]{.45\textwidth}
         \centering
         \includegraphics[width=\linewidth]{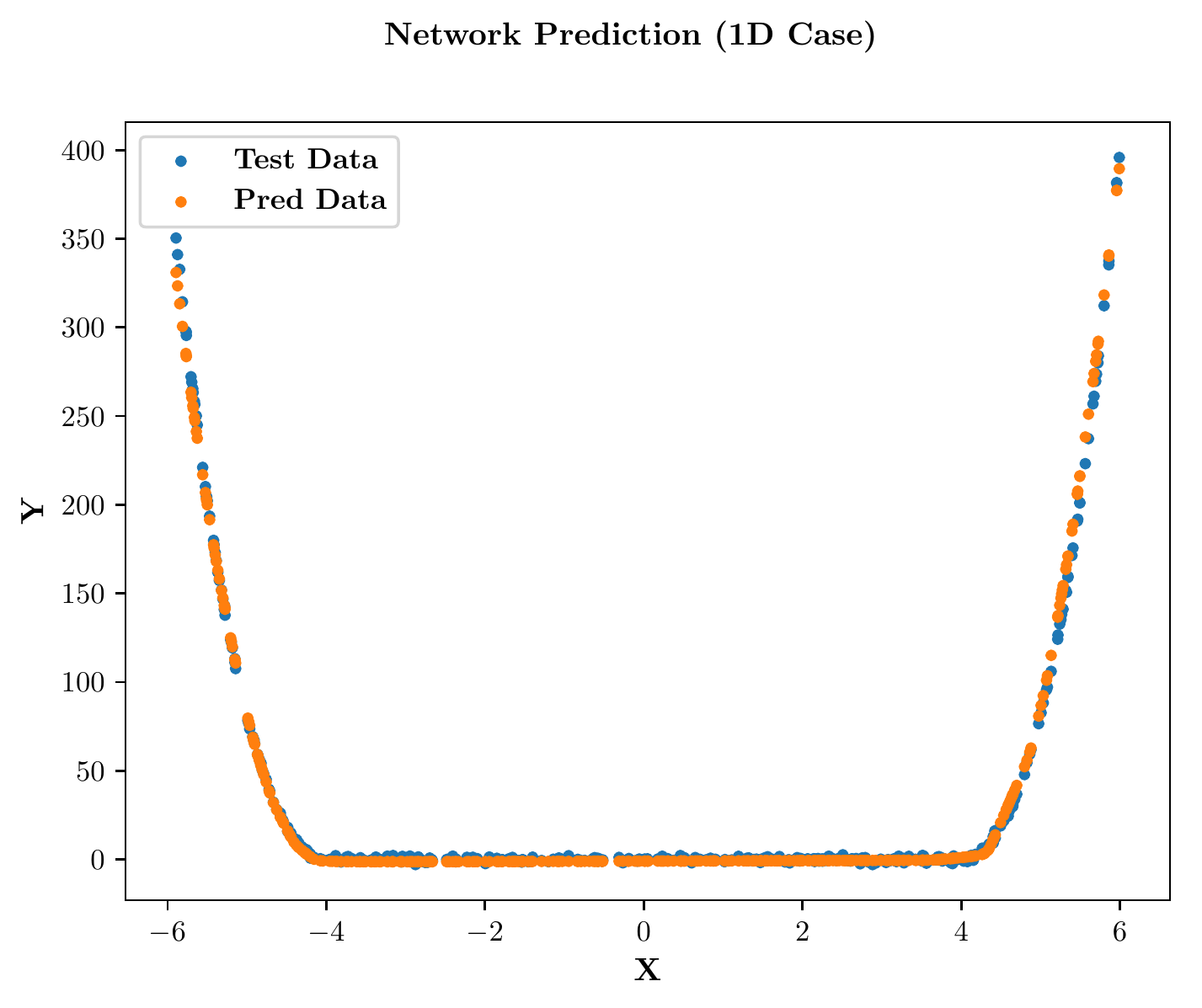}
         \caption{$60\%$ or more of function $\phi_2$}
     \end{subfigure}
        \caption{Plot of Test and Predicted Data for the functions $\phi_1$ and $\phi_2$ for 1-dimensional test case.}
        \label{fig:1Dmodel}
\end{figure}
\par
As seen in figure \ref{fig:1Dmodel}, it is clear that the neural network was able to approximate the given functions by reducing the loss function to the minimum. 
\\
\par
As discussed in Section \ref{section:ProblemSetting}, to set up a multi-valued data set  we combine fraction of both the sets $\Phi_1$ and  $\Phi_2$ respectively, 
to form a new data set $\Phi$ as per our requirement. The two data sets were combined in different fractions, trained using our neural network and then tested on the test data which is $20\%$ of the new combined data. The noise was added to the data set to replicate the real-world data. The network was trained using logcosh loss function to examine the network behaviour.
%how the network behaves when clusters of data exists. 
%Schon oft gesagt: By using logcosh loss function, we aim to classify the clusters of data.
To compare the functionality of different loss functions, the network was also trained with MSE and MAE loss function using a similar setting.
\\
\par
\begin{figure}[ht!]
    \centering
    \includegraphics[width=5.5cm]{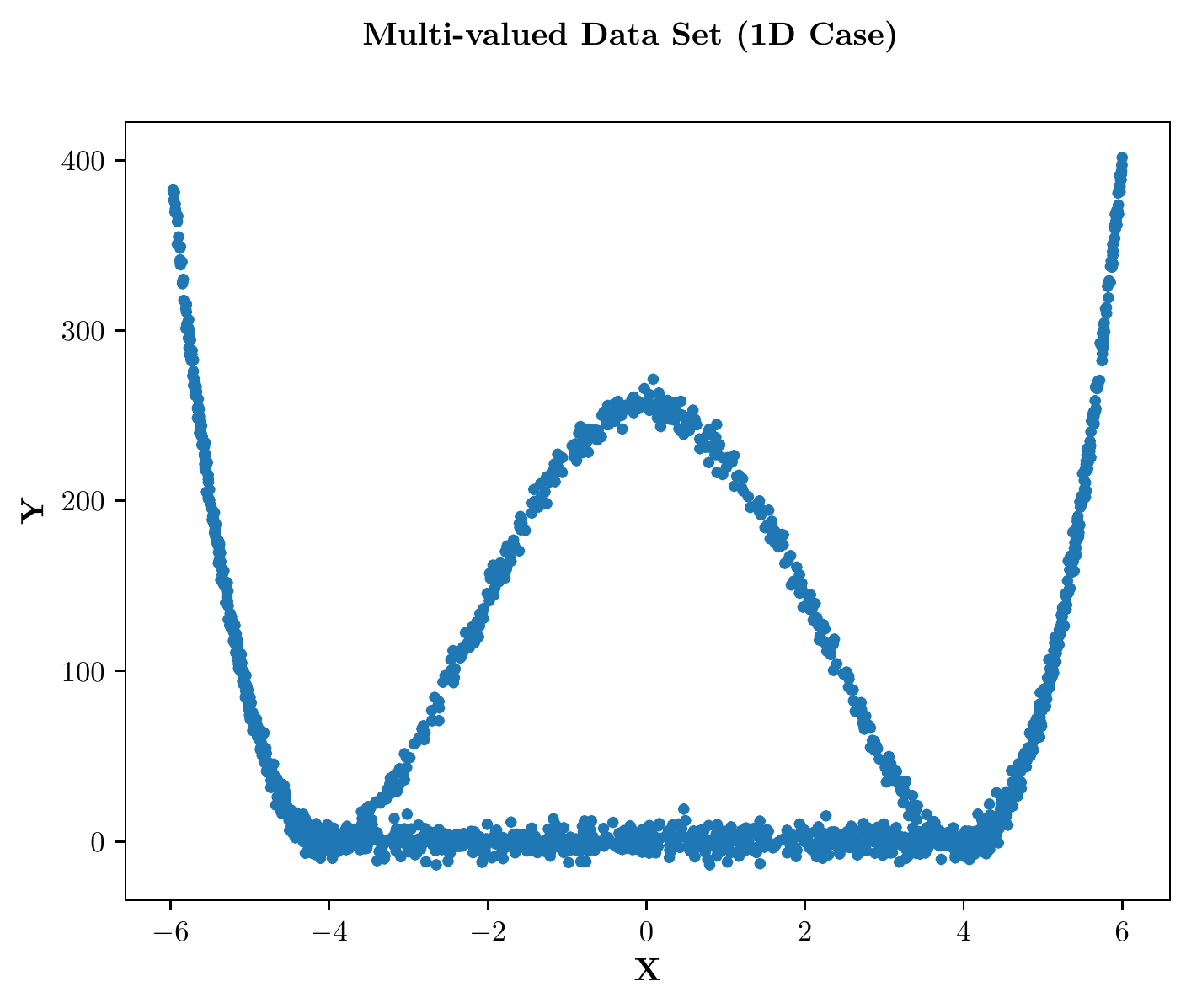}
    \caption{Plot of data set with noise for 1-dimensional case}
    \label{fig:modelmixed}
\end{figure}

The combined function %data set 
can be written as follows :
\begin{align}
\Phi (\vek x) = \left\{ \begin{array}{cc} 
                \Phi_1 \\ %& % \hspace{5mm} x \in [-6,-4) \\
                %\Hat{\phi} & \hspace{5mm} x \in [-4,4] \\
                \Phi_2 % & \hspace{5mm} x \in (4,6] \\
                \end{array} \right\}.
\end{align}
where $\Phi(\vek x)$ is a combination of both multi-valued (the multi-valued region where each $\vek x$ has two possible outputs $ y$ as shown in the figure~\ref{fig:modelmixed} and single 
valued function (defined on $[-6,-4)\bigcup(4,6]$).   %and $\Hat{\phi}$ represents . 

\subsubsection{Network behaviour}
\begin{figure}[ht!]
     \centering
     \begin{subfigure}[b]{.45\textwidth}
         \centering
         \includegraphics[width=\linewidth]{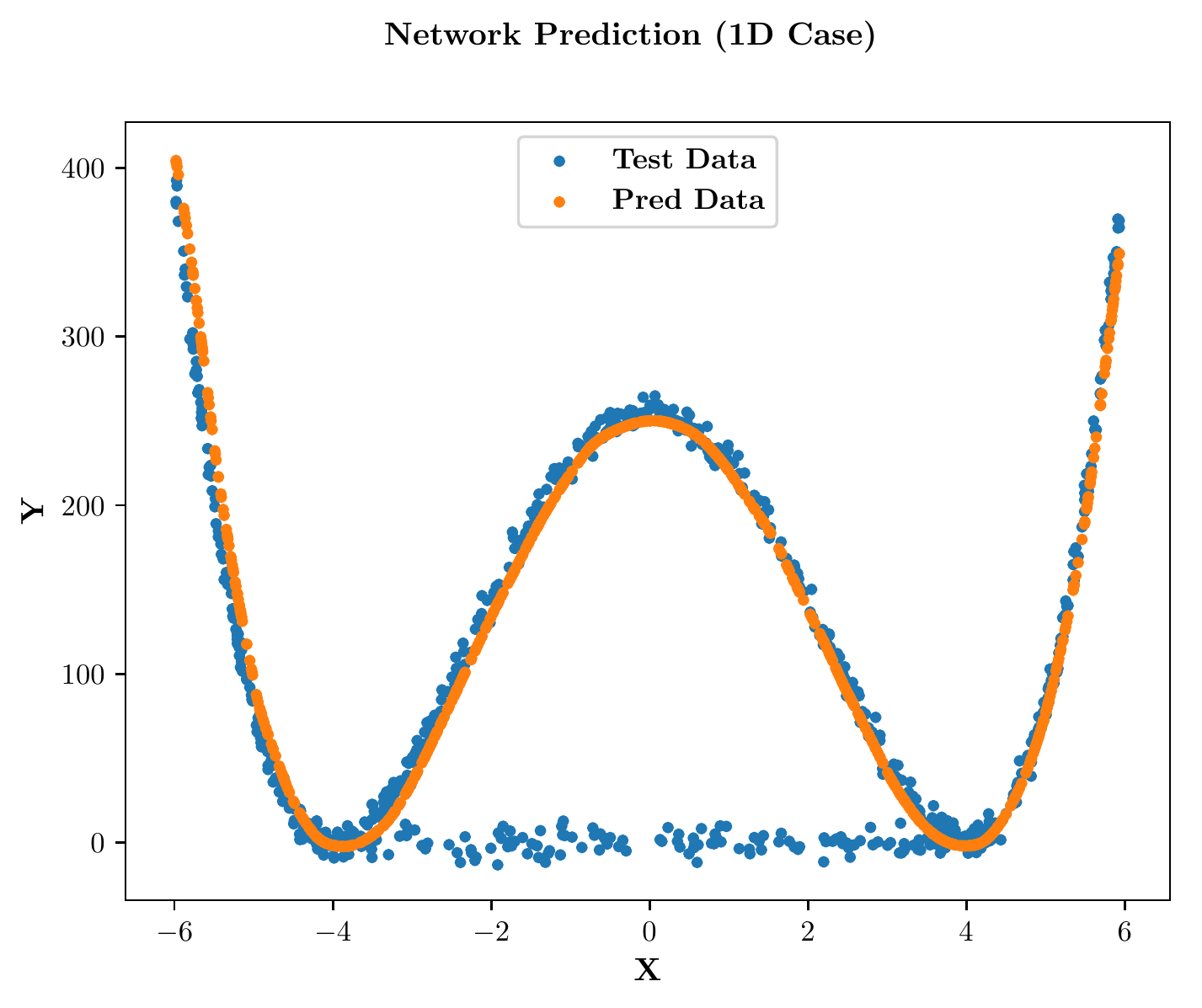}
         \caption{$60\%$ or more of function $\Phi_1$}
     \end{subfigure}
     \begin{subfigure}[b]{.45\textwidth}
         \centering
         \includegraphics[width=\linewidth]{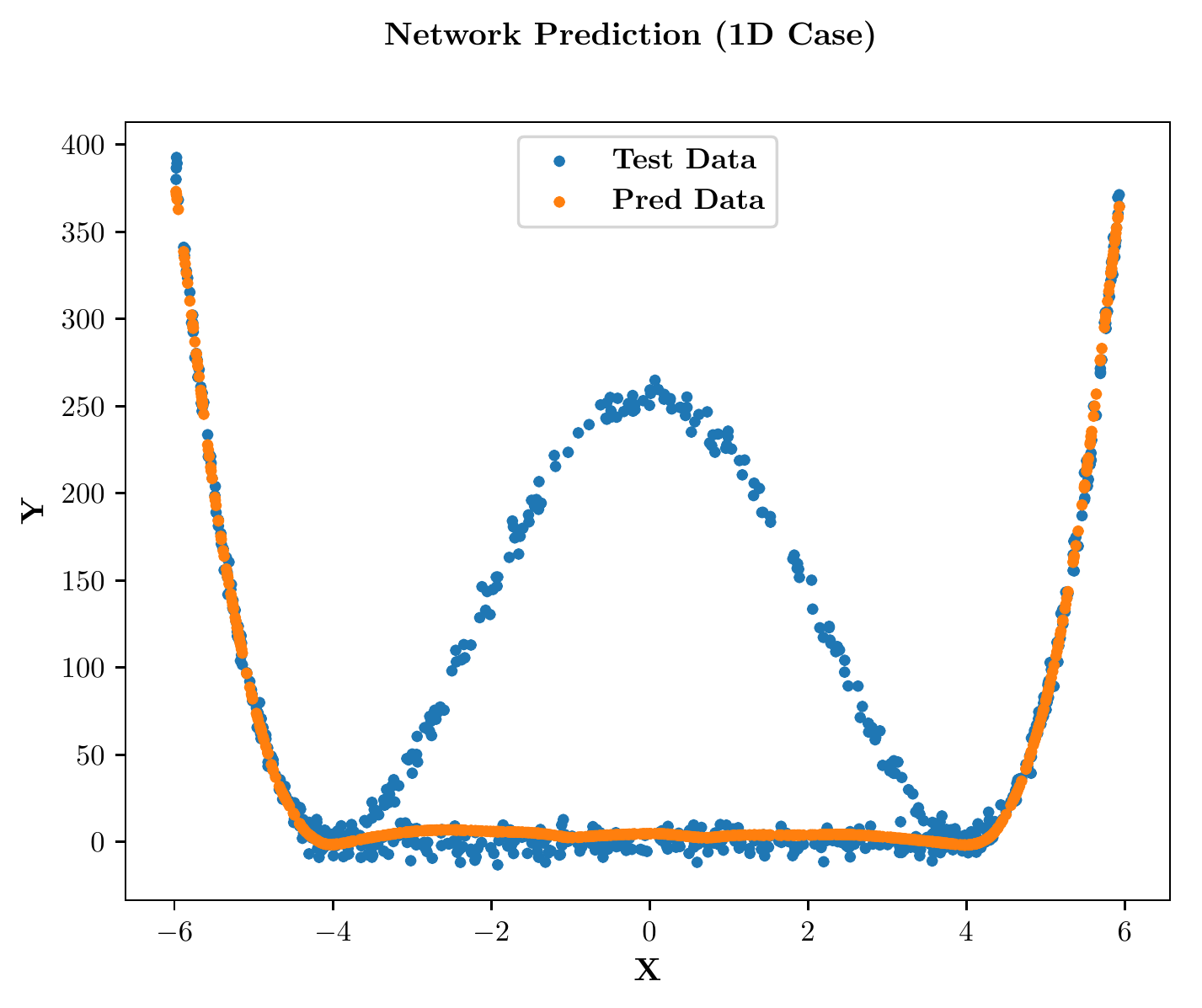}
         \caption{$60\%$ or more of function $\Phi_2$}
     \end{subfigure}
        \caption{Plot of Test and Predicted Data for the mixed data set in the 1-dimensional case. The plots illustrate the behaviour of the network when tested on the test data. The network accurately predicts one of the 2 functions depending on the fraction of functions considered when trained using log-cosh loss function.}
        \label{fig:netowrkbehavious1D}
\end{figure}

\begin{figure}[ht!]
     \centering
     \begin{subfigure}[b]{.45\textwidth}
         \centering
         \includegraphics[width=\linewidth]{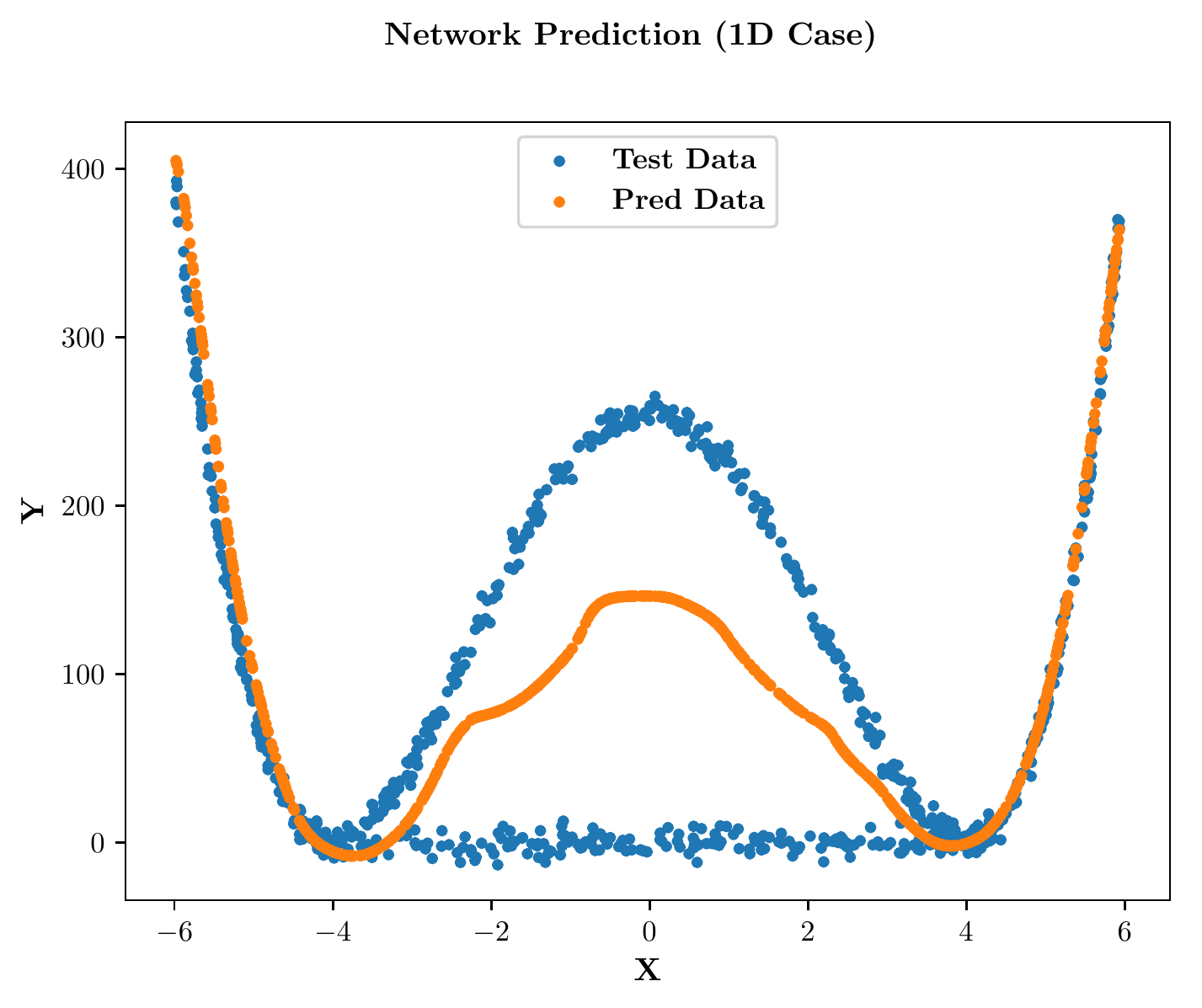}
         \caption{Network behaviour when the model was trained with MSE loss function. The network predicts the weighted mean of the 2 functions depending on the fractional composition\\}
     \end{subfigure}
     \begin{subfigure}[b]{.45\textwidth}
         \centering
         \includegraphics[width=\linewidth]{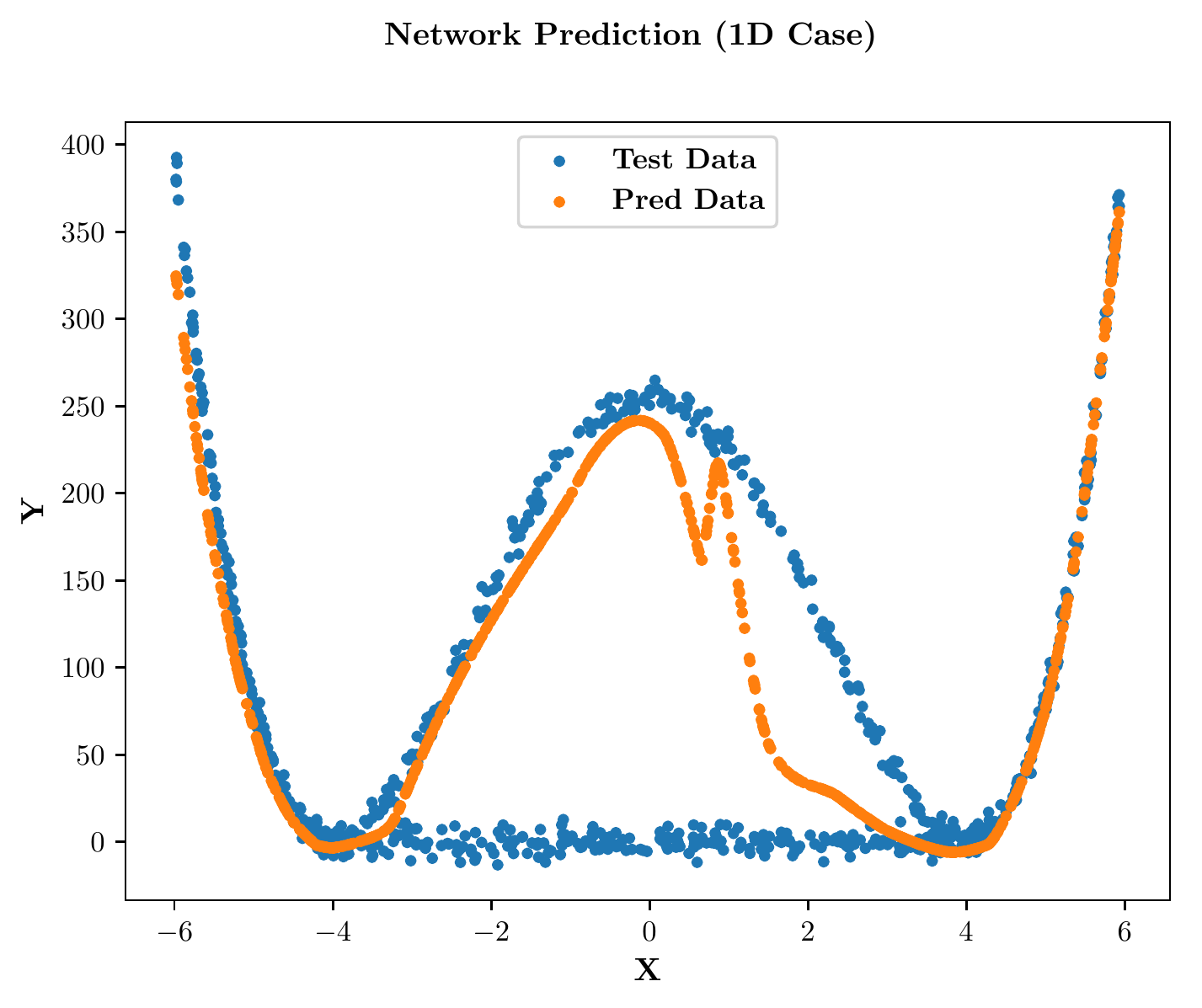}
         \caption{Network behaviour when the model was trained with MAE loss function. Though the network predicts one of the 2 clusters often, the prediction oscillates between the two when the fractions of the clusters are nearly equal.}
     \end{subfigure}
        \caption{Plot of Test and Predicted Data for the mixed data set in the 1-dimensional case when the network is trained using (a) MSE and (b) MAE loss function.}
        \label{fig:netowrkbehaviour1D_differentloss}
\end{figure}

In this section, the behaviour of our network based on the chosen network architecture is discussed. As discussed earlier the network was trained with a different fraction of the two chosen functions and then tested on the test data. The network was completely trained using log-cosh loss function. It can be seen that, when using the log-cosh loss, the network predicted one of the two chosen function with high accuracy and not the mean of the two functions. The network predicted the function $\Phi_1$, when $60\%$ or more of the function $\Phi_1$-generated data was chosen in the combined data set and it predicted the function $\Phi_2$ otherwise, as shown in the figure~\ref{fig:netowrkbehavious1D}.
%It is clear that 
As could be expected (see Section  \ref{section:logcosh}),  the logcosh loss function learned the bigger cluster of data, unlike the mean square error loss which  learned the mean of the two functions or the absolute error which would oscillate between the two chosen functions as shown in Figure \ref{fig:netowrkbehaviour1D_differentloss}. As mentioned in section \ref{section:logcosh}, the MSE loss function gets affected by the minor cluster due to squaring and thus finds the weighted mean between the two depending on composition of the cluster. Unlike MSE, MAE functions similar to that of logcosh and tries to find one of the two clusters. However, since it is non smooth and has non-continuous derivative, the prediction oscillates between the clusters when the composition of the clusters are nearly equal. 

\subsection{Two-Dimensional Test Case}
\subsubsection{Test Problem}
We now choose a 2-dimensional case based on the concept discussed in section \ref{section:ProblemSetting}. Similar to the 1D case, two 2 dimensional single-valued functions were combined in different fractions to form the multi-valued data set $\Phi$, learned by the neural network and finally, the behaviour of our network based on these data sets was analysed.

%Two simple 2D functions were chosen to validate our claim.
The two functions 
\begin{equation}
f_{1}( x, y) = xy(2x+2y)
\end{equation}
\begin{equation}
f_{2}( x,  y) = xy(x^2+y^2)
\end{equation}
were  used as arguments to the sigmoid function. The main reason to use the sigmoid function was to keep the range between (0,1).
\begin{equation}
\Phi_{1}( x,  y) = \text{sigmoid}(f_{1}( x,  y)) = \frac{1}{1+e^{-f_1( x,  y)}}
\end{equation}
\begin{equation}
\Phi_{2}( x,  y) = \text{sigmoid}(f_{2}( x,  y)) = \frac{1}{1+e^{-f_2( x,  y)}}
\end{equation}
To set up a multi-valued data set we combined both the data sets $ \Phi_1$ and  $ \Phi_2$  of the above functions in different fractions to form a combined data set $ \Phi$ as per our requirement. Noise was added to the data set to replicate the real-world scenario. 

\subsubsection{Training Strategy}
The neural network was trained with this data set and then predicted on the test data which is $20\%$ of the total combined data.
\begin{figure}[ht]
    \centering
    \includegraphics[width=10cm]{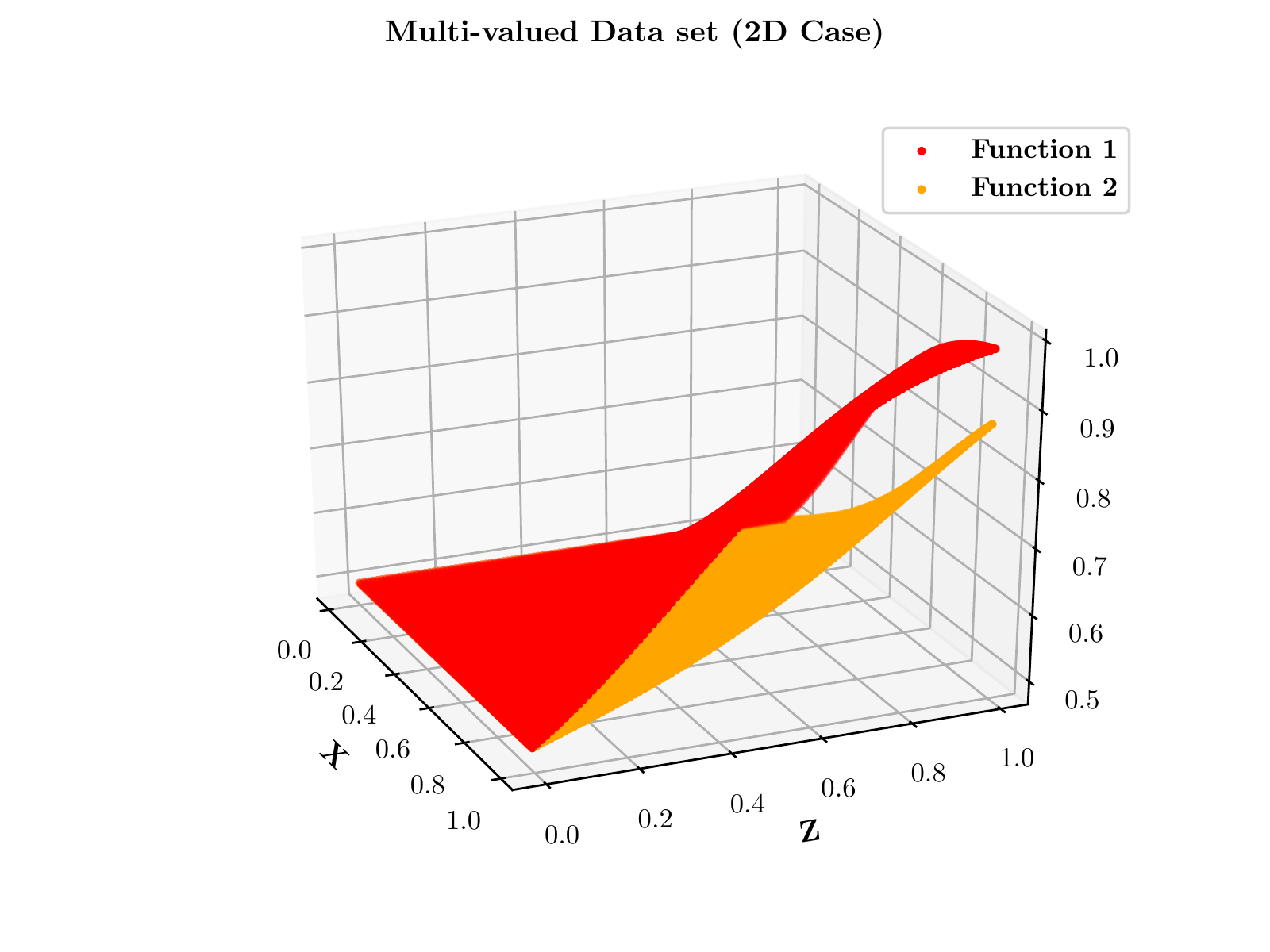}
    \caption{Plot of data set without noise for a 2-dimensional data set.}
    \label{fig:2Dmixed}
\end{figure}
Figure~\ref{fig:2Dmixed} shows the plot of the combined data set without noise, where red and orange represent the function $\Phi_1$ and function $\Phi_2$ respectively. As discussed earlier, in this case, for  given nearby $( x, y)_1$ and $( x, y)_2$, we have two distant values $ z_1$ and $ z_2$ despite being very close to each other. The network with logcosh loss function is trained with different fractions of the sets $\Phi_1$ and $\Phi_2$ joined into $\Phi$ in an aim to classify the two. % based on the weight given to the functions.

\subsubsection{ Network behaviour}
A very noisy data set was used to train the network -- the 2 populations cannot be easily distinguished by visualization. After training the network with a combination $\Phi$ of different fractions of $\Phi_1$ and $\Phi_2$, similar to the 1D case, a clear rule was visible when the logcosh loss function was used. 
\begin{figure}[ht!]
     \centering
     \begin{subfigure}[b]{.45\textwidth}
         \centering
         \includegraphics[width=\linewidth]{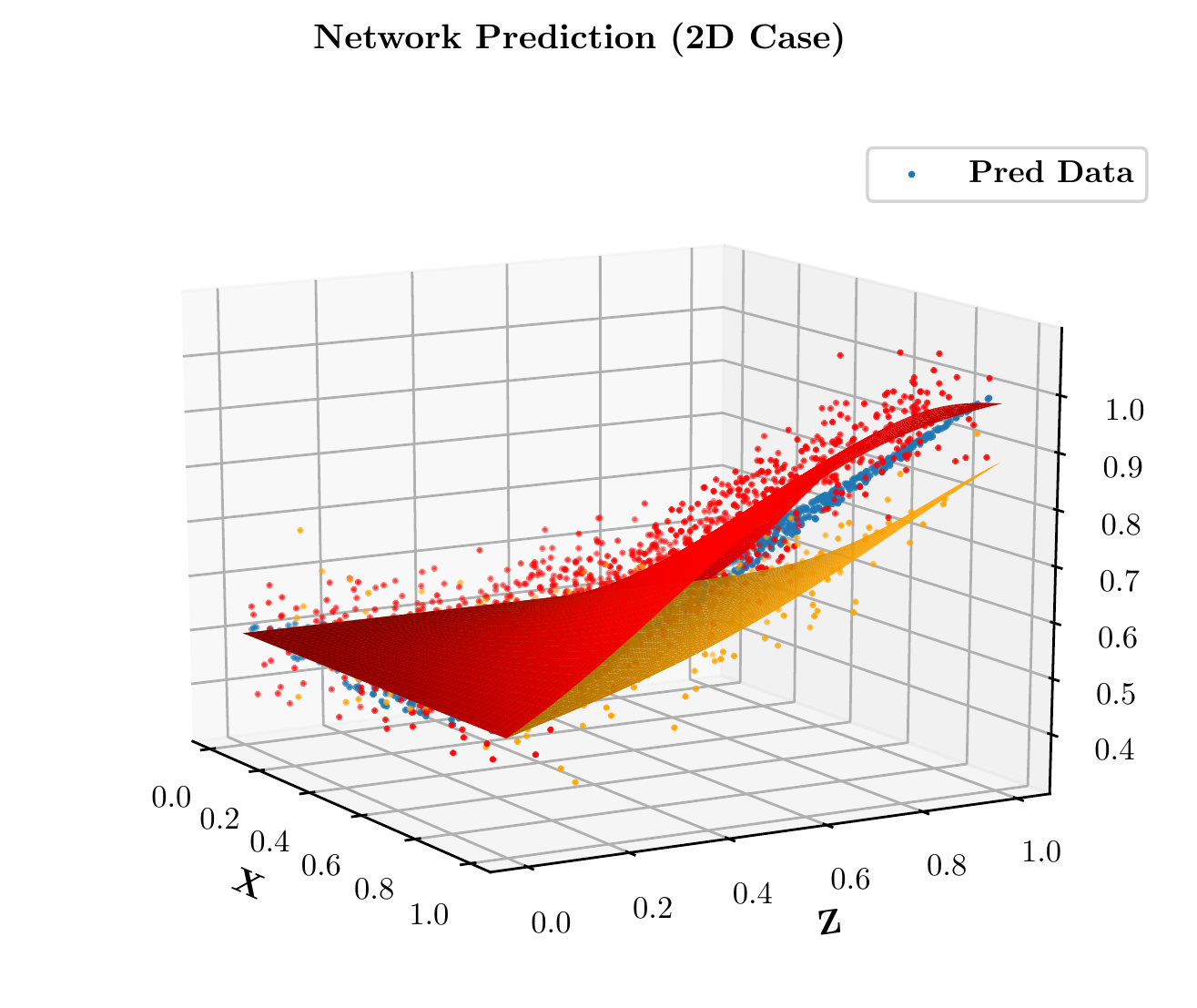}
         \caption{$60\%$ or more of function $\Phi_1$ values in data}
     \end{subfigure}
     \begin{subfigure}[b]{.45\textwidth}
         \centering
         \includegraphics[width=\linewidth]{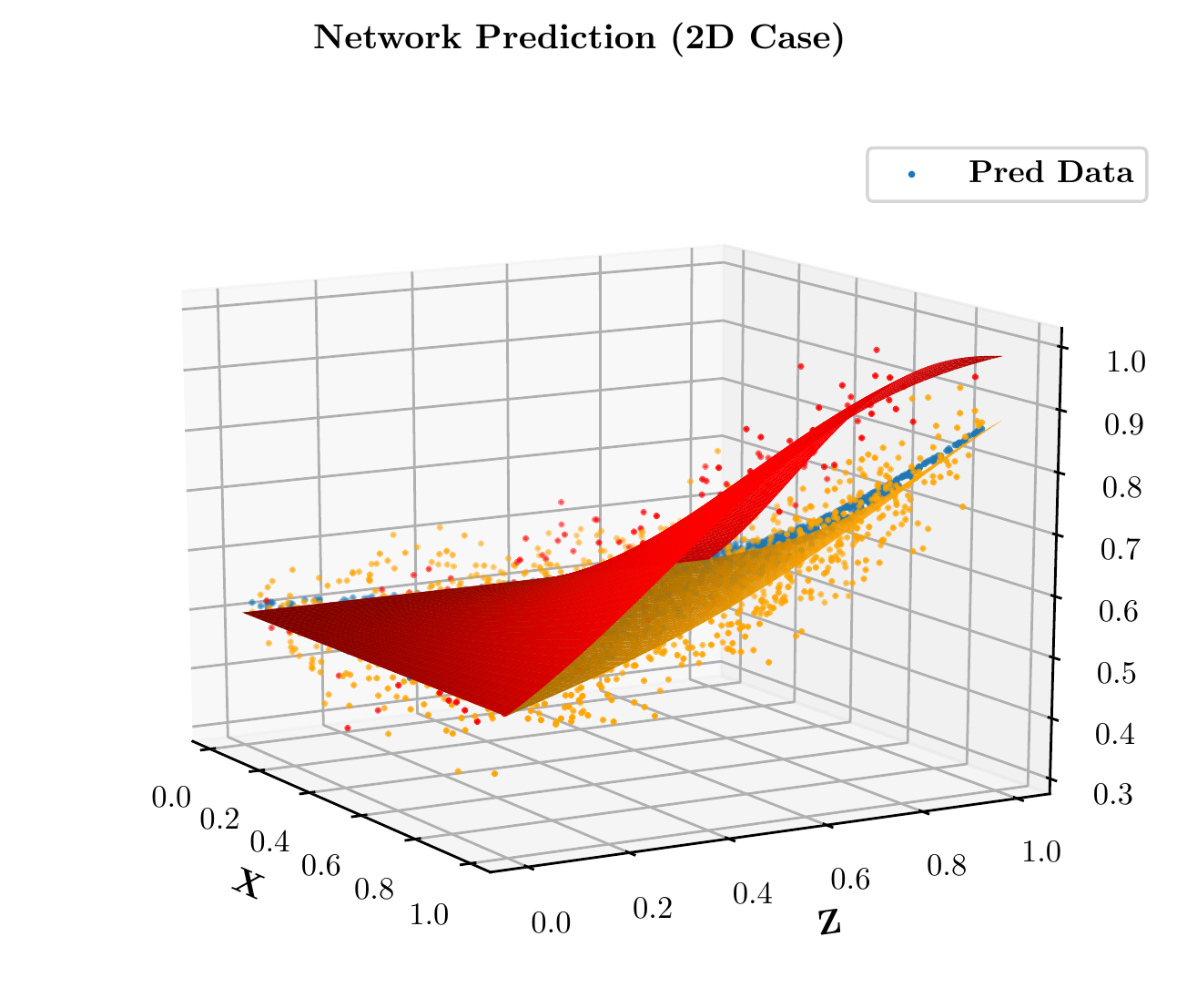}
         \caption{$60\%$ or more of function $\Phi_2$ values in data}
     \end{subfigure}
        \caption{Plot of Test and Predicted Data for the mixed data set in the 2-dimensional case. The network accurately predicts one of the two functions  depending on the fractional composition of the data set.}
        \label{fig:2Dpredicted}
\end{figure}
The network predicted the function $\Phi_1$ when $60\%$ or more of $\Phi$ consisted values of  $\Phi_1$ %, was chosen in the fraction of the combined data set 
and  $\Phi_2$ when $60\%$ or more of $\Phi$ consisted values of  $\Phi_2$ as shown in the figure~\ref{fig:2Dpredicted}. In the plots, the red scatter points represent the function   $\Phi_1$ with noise and the red surface plot represents the values $ \Phi_1$ of function ${\Phi}$ without noise. Similarly, for the function $\Phi_2$, orange scatter points and orange surface plot represents the function with and without noise respectively. Finally, the blue scatter points represent the predicted value. The functions were plotted without noise for better visualisation. From figure \ref{fig:2Dpredicted}, it is clear that the network learnt one of the 2 functions accurately without being influenced by noise. It can be therefore confirmed that the neural network predicts the bigger of the two clusters when logcosh loss function is used.

%network predicts with noise

\section{Conclusion}
Based on the network behaviour, we claim that a network with logcosh loss function can be used to classify the data when clusters of data exist. It can be concluded that in case of clustered data, an artificial neural network with logcosh learns the bigger cluster rather than the mean of the two. Even more so, the ANN when used for regression of a set-valued function, will learn a value close to one of the choices, in other words, one branch of the set-valued function, while a mean-square-error NN will learn the value in between. Based on the above result we have a neural network that not only helps in classifying the data based on the invisible features but also predicts the majority cluster with high accuracy. In the real world scenario, the unavailability of enough parameters to build the regression model is always a major problem and therefore it becomes increasingly difficult to represent the model based on the available limited data. Using this theory, we can classify the clusters of data based on an invisible feature which is not available to us beforehand. It can be also used to validate if there are enough features to represent the model. In other words, we can confirm if a feature is essential to represent the model.

%\input{shortupdate.tex}
%\input{nn.tex}
%\appendix
%\input{appendix.tex}

%\bibliographystyle{plain}%/usr/share/texlive/texmf-dist/bibtex/bst/base/apalike.bst}
%bibliography{../../matthies/sizeEffectsFailure/plasticity.bib,./update.bib}

\printbibliography[heading=bibintoc]

@book{clustering, title={Data Clustering}, url={https://www.oreilly.com/library/view/data-clustering/9781466558229/}, journal={O'Reilly Online Learning}, publisher={Chapman and Hall/CRC}, author={Aggarwal, Charu C. and Reddy, Chandan K.}
}

@BOOK{1999kats.book.....S,
       author = {{Stuart}, Alan and {Ord}, J. Keith and {Arnold}, Steven},
        title = "{Kendall's advanced theory of statistics. Vol.2A: Classical inference and the linear model}",
         year = 1999,
       adsurl = {https://ui.adsabs.harvard.edu/abs/1999kats.book.....S}
}

@book{Goodfellow-et-al-2016,
    title={Deep Learning},
    author={Ian Goodfellow and Yoshua Bengio and Aaron Courville},
    publisher={MIT Press},
    note={\url{http://www.deeplearningbook.org}},
    year={2016}
}

@article{loss,
author = {Nie, Feiping and Zhanxuan, Hu and Li, Xuelong},
year = {2018},
month = {01},
pages = {37-52},
title = {An investigation for loss functions widely used in machine learning},
volume = {18},
journal = {Communications in Information and Systems},
doi = {10.4310/CIS.2018.v18.n1.a2}
}

@online{lossfuncations,
    Author = {Prince Grover},
    Title = {5 Regression Loss Functions All Machine Learners Should Know},
    Url = {https://heartbeat.fritz.ai/5-regression-loss-functions-all-machine-learners-should-know-4fb140e9d4b0},
    Year = {2018}}

@inbook{doi:https://doi.org/10.1002/9781118445112.stat06938,
author = {Kaufmann, Jörg and Schering, AG},
publisher = {American Cancer Society},
isbn = {9781118445112},
title = {Analysis of Variance ANOVA},
booktitle = {Wiley StatsRef: Statistics Reference Online},
chapter = {},
pages = {},
doi = {https://doi.org/10.1002/9781118445112.stat06938},
url = {https://onlinelibrary.wiley.com/doi/abs/10.1002/9781118445112.stat06938},
eprint = {https://onlinelibrary.wiley.com/doi/pdf/10.1002/9781118445112.stat06938},
year = {2014}
}

\end{document}